\documentclass{article}

\usepackage{amsmath,amsfonts,bm}

\def\eqref#1{equation~\ref{#1}}

\def\1{\bm{1}}

\DeclareMathAlphabet{\mathsfit}{\encodingdefault}{\sfdefault}{m}{sl}
\SetMathAlphabet{\mathsfit}{bold}{\encodingdefault}{\sfdefault}{bx}{n}

\newcommand{\E}{\mathbb{E}}

\usepackage{floatrow}
\newfloatcommand{capbtabbox}{table}[][\FBwidth]

\usepackage{microtype}
\usepackage{graphicx}
\usepackage{booktabs} 
\usepackage{xcolor}

\definecolor{dgreen}{rgb}{0.00,0.49,0.00}
\definecolor{dblue}{rgb}{0,0.08,0.75}
\RequirePackage[colorlinks,citecolor=dgreen,urlcolor=dblue,linkcolor=dblue]{hyperref}

\usepackage[utf8]{inputenc} 
\usepackage[T1]{fontenc}    
\usepackage{url}            
\usepackage{booktabs}       
\usepackage{amsfonts}       
\usepackage{nicefrac}       
\usepackage{microtype}      
\usepackage{graphicx}
\usepackage{subcaption}

\usepackage[font={footnotesize}]{caption}
\usepackage{amssymb}
\usepackage{amsmath}
\usepackage{amsthm}
\usepackage{tabularx}

\renewcommand{\paragraph}[1]{{\bf #1.}}

\newcommand{\supp}{\textnormal{supp}}

\newcommand{\sailb}{SAIL-b}
\newcommand{\gailb}{GAIL-b}

\makeatletter
\newcommand\footnoteref[1]{\protected@xdef\@thefnmark{\ref{#1}}\@footnotemark}
\makeatother

\usepackage[tableposition = top]{caption}

\usepackage{natbib}

\usepackage{algorithm}
\usepackage{algorithmic}

\usepackage[nameinlink,capitalize]{cleveref}

\usepackage{thmtools, thm-restate}
\declaretheorem[name=Theorem,refname=Thm.]{theorem}

\declaretheorem[name=Proposition,refname=Prop.,sibling=theorem]{proposition}

\crefname{assumption}{assumption}{assumptions}

\crefname{equation}{Eq.}{Eqs.}
\crefname{figure}{Fig.}{Figs.}

\newcommand{\pp}{{\mathbb{P}}}

\usepackage{enumitem}

\renewcommand{\paragraph}[1]{~\newline{\bfseries #1.}}

\title{Support-weighted Adversarial Imitation Learning}

\date{}

\author{Ruohan Wang$^{1}$ \quad Carlo Ciliberto$^1$ \quad Pierluigi Amadori$^{1}$ \quad Yiannis  Demiris$^{1}$\\ 
{\small $^1$ Department of Electrical and
Electronic Engineering, Imperial College London, SW7 2BT, United Kingdom}\\
{\footnotesize\tt \{r.wang16,c.ciliberto,p.amadori,y.demiris\}@imperial.ac.uk} }

\begin{document}

\maketitle

\begin{abstract}
\noindent Adversarial Imitation Learning (AIL) is a broad family of imitation learning methods designed to mimic expert behaviors from demonstrations.
While AIL has shown state-of-the-art performance on imitation learning with only small number of demonstrations, it faces several practical challenges such as potential training instability and implicit reward bias.
To address the challenges, we propose Support-weighted Adversarial Imitation Learning (SAIL), a general framework that extends a given AIL algorithm with information derived from support estimation of the expert policies.
SAIL improves the quality of the reinforcement signals by weighing the adversarial reward with a confidence score from support estimation of the expert policy.
We also show that SAIL is always at least as efficient as the underlying AIL algorithm that SAIL uses for learning the adversarial reward.
Empirically, we show that the proposed method achieves better performance and training stability than baseline methods on a wide range of benchmark control tasks.
\end{abstract}

\section{Introduction}
Adversarial Imitation Learning (AIL) algorithms is a powerful class of methods that learn to imitate expert behaviors from demonstrations. AIL alternates between learning a reward function via adversarial training, and performing reinforcement learning (RL) with the learned reward function. AIL has been shown to be effective with only a small of expert trajectories, with no further access to other RL signals. AIL is also more robust and mitigates the issue of distributional drift from behavioral cloning~\cite{ross2011reduction}, a classical imitation learning method that requires large amount of training data to generalize well. Generative Adversarial Imitation Learning (GAIL)~\citep{ho2016generative}, an early and influential AIL method, shows the equivalence between inverse reinforcement learning settings and Generative Adversarial Networks (GANs)~\cite{goodfellow2014generative}. This observation motivates casting imitation learning as a distribution matching task between the expert and the RL agent policies. Recent works have sought to improve various aspects of AIL, such as robustness to changes in environment dynamics~\cite{fu2017learning}, and improving sample efficiency of environment interactions~\cite{nagabandi2018neural}. However, AIL still faces several practical challenges associated with adversarial training, including potential training instability~\cite{salimans2016improved, brock2018large}, and implicit reward bias~\cite{kostrikov2018discriminator}.

\citealt{wang2019red} demonstrated that imitation learning is also feasible by constructing a fixed reward function via support estimation of the expert policy. Since support estimation only requires expert demonstrations, the method sidesteps the training instability associated with adversarial training. However, we show in \cref{sec:mujoco} that the reward learned via support estimation deteriorates and leads to poor performance when the expert data is sparse.

Support estimation and adversarial reward represent two different yet complementary RL signals for imitation learning, both learnable from expert demonstrations. We unify both signals into \textit{Support-weighted Adversarial Imitation Learning} (SAIL), a general framework that weighs the adversarial reward with a confidence score derived from support estimation of the expert policy. Consequently, SAIL improves the quality of the learned reward to address potential training instability and reward bias. We highlight that SAIL may be efficiently applied on top of many existing AIL algorithms such as GAIL and Discriminator-Actor-Critic~\cite{kostrikov2018discriminator}. In addition, it can be shown that SAIL is at least as efficient as the underlying AIL method that SAIL leverages for learning the adversarial reward. In the experiments on a series of benchmark control tasks, we demonstrate that SAIL achieves better performance and training stability, as well as mitigates the implicit reward bias.

Our main contribution includes:
\begin{itemize}[partopsep=0em,itemsep=0em,topsep=0em]
    \item We propose a general framework for adversarial imitation learning, combining both adversarial rewards and support estimation of the expert policy.
    \item The proposed method is easy to implement, and may be applied to many existing AIL algorithms.
    \item We show SAIL improves performance and training stability, and better mitigates reward bias over the baseline methods.
\end{itemize}

The rest of the paper is organized as follows: we review the relevant background and literature in \cref{sec:bg}. In \cref{sec:method}, we detail the proposed method and the theoretical analysis on its sample efficiency with respect to the expert data. We present the experiment results in \cref{sec:exp} and conclude in \cref{sec:con}.

\section{Background}
\label{sec:bg}
We recall the definition of Markov Decision Process and introduce formal notations used in this work. We also review the related literature on imitation learning.

\subsection{Task Setting}
We consider an infinite-horizon discounted {\itshape Markov Decision Process (MDP)} $(S, A, P, r, p_0, \gamma)$, where $S$ is the set of states, $A$ the set of actions, $P :
S \times A \times S \rightarrow [0, 1]$ the transition probability, $r : S \times A \rightarrow \mathbb{R} $ the reward function, $p_0 : S \rightarrow [0, 1]$ the distribution over initial states, and $\gamma \in (0, 1)$ the discount factor. Let $\pi$ be a stochastic policy $\pi : S \times A \rightarrow [0, 1]$ with expected discounted reward $\E_{\pi}(r(s, a)) \triangleq \E(\sum_{t=0}^{\infty} \gamma^tr(s_t, a_t))$ where $s_0 \sim p_0$, $a_t \sim \pi(\cdot|s_t)$, and $s_{t+1} \sim P(\cdot |s_t, a_t)$ for $t \geq 0$. We denote $\pi_E$ the expert policy.

\subsection{Imitation Learning}

Behavioral Cloning (BC) learns a policy $\pi : S \rightarrow A$ directly from expert trajectories via supervised learning. BC is simple to implement, and effective when expert data is abundant. However, BC is prone to distributional drift: the state distribution of expert demonstrations deviates from that of the agent policy, due to accumulation of small mistakes during policy execution. Distributional drift may lead to catastrophic errors~\cite{ross2011reduction}. While several methods address the issue \cite{ross2010efficient, sun2017deeply}, they often assume further access to the expert during training.

Inverse Reinforcement Learning (IRL) first estimates a reward from expert demonstrations, followed by RL using the estimated reward~\cite{ng2000algorithms, abbeel2004apprenticeship}. Building upon a maximum entropy formulation of IRL~\cite{ziebart2008maximum}, \citealt{finn2016connection} and \citealt{fu2017learning} explore adversarial IRL and its connection to Generative Adversarial Imitation Learning~\cite{ho2016generative}.

\subsubsection{Adversarial Imitation Learning}
Generative Adversarial Imitation Learning (GAIL)~\cite{ho2016generative} is an early and influential work on AIL. It casts imitation learning as distribution matching between the expert and the RL agent. Specifically, the authors show the connection between IRL and GANs and formulate the following a minimax game:
\begin{equation}
    \min_{\pi} \max_{D \in (0, 1)} \E_{\pi} (\log D(s, a)) + \E_{\pi_E}(\log (1-D(s, a))),
\label{eq:gail}
\end{equation}
where the expectations $\E_{\pi}$ and $\E_{\pi_E}$ denote the joint distributions over state-actions of the RL agent and the expert, respectively. GAIL is able to achieve expert performance with a small number of expert trajectories on various benchmark tasks. However, GAIL is relatively sample inefficient with respect to environment interaction, and inherits issues associated with adversarial training, such as vanishing gradients, training instability and overfitting to expert demonstrations~\cite{arjovsky2017towards, brock2018large}.

GAIL has inspired many follow-up works aimed at improving the efficiency and stability of AIL methods. For instance, Generative Moment Matching Imitation Learning~\cite{kim2018imitation} replaces the adversarial reward with a non-parametric maximum mean discrepancy estimator to sidestep adversarial learning. \citealt{baram2017end} improve sample efficiency with a model-based RL algorithm. In addition, \citealt{kostrikov2018discriminator} and \citealt{sasaki2018sample} demonstrate significant gain in sample efficiency with off-policy RL algorithms. Furthermore, Generative Predecessor Models for Imitation Learning~\cite{schroecker2019generative} imitates the expert policy using generative models to reason about alternative histories of demonstrated states.

The proposed method extends the broad family of AIL algorithms additional information. In particular, we improve the quality of the learned reward by weighing the adversarial reward with a score derived from support estimation of the expert policy. The proposed method is therefore complementary and orthogonal to many aforementioned techniques for improving the algorithmic efficiency and stability.

\subsubsection{Imitation Learning via Support Estimation} Alternative to AIL, \citealt{wang2019red} demonstrate the feasibility of using a fixed RL reward via estimating the support of the expert policy from expert demonstrations. Connecting kernel-based support estimation~\cite{de2014universally} to Random Network Distillation~\cite{burda2018exploration}, the authors propose Random Expert Distillation (RED) to learn a reward function based on support estimation. Specifically, RED learns the reward parameter $\hat{\theta}$ by minimizing:
\begin{equation}
\label{eq:red}
    \min_{\hat{\theta}}\E_{s,a \sim \pi_E} ||f_{\hat{\theta}}(s, a) - f_{\theta}(s, a)||_2^2,
\end{equation}
where $f_{\theta} : S\times A \rightarrow \mathbb{R}^K$ projects $(s,a)$ from expert demonstrations to some embedding of size $K$, with randomly initialized $\theta$. The reward is then defined as:
\begin{equation}
    r_{red}(s, a) = \exp(-\sigma||f_{\hat{\theta}}(s, a) - f_{\theta}(s, a)||_2^2),
\end{equation}
where $\sigma$ is a hyperparameter. As optimizing \cref{eq:red} only requires expert data, RED sidesteps adversarial learning, and casts imitation learning as a standard RL task using the learned reward. While RED works well given sufficient expert data, we show in the experiments that its performance suffers in the more challenging setting of sparse expert data.

\section{Method}
\label{sec:method}
Formally, we consider the task of learning a reward function $\hat r(s, a)$ from a finite set of trajectories $\{\tau_i\}_{i=1}^N$, sampled from the expert policy $\pi_E$ within a MDP. Each trajectory is a sequence of state-action tuples in the form of $\tau_i=\{s_1, a_1, s_2, a_2, ..., s_T, a_T\}$. Assuming that the expert trajectories are consistent with some latent reward function $r^*(s, a)$, we aim to learn a policy that achieves good performance with respect to $r^*(s, a)$ by applying RL on the learned reward function $\hat r(s, a)$.

In this section, we first discuss the advantages and shortcomings of AIL to motivate our method. We then introduce {\itshape Support-weighted Adversarial Learning} (SAIL), and present a theoretical analysis that compares SAIL with the underlying AIL method that SAIL uses for adversarial reward learning. In particular, we consider GAIL for adversarial reward learning.

\subsection{Adversarial Imitation Learning}
\label{sec:gail}
A clear advantage of AIL resides in its low sample complexity with respect to expert data. For instance, GAIL requires as little as 200 state-action tuples from the expert to achieve imitation. The reason is that the adversarial reward may be interpreted as an effective exploration mechanism for the RL agent. To see this, consider the learned reward function under the optimality assumption. With the optimal discriminator to \cref{eq:gail} $D^*(s, a) = \frac{p_{\pi}(s, a)}{p_{\pi_{E}}(s, a)+p_{\pi}(s, a)}$, a common reward for GAIL is
\begin{equation}
\begin{split}
    r_{gail}(s, a) & = -\log(D^*(s, a)) \\
& = \log\left(1+\frac{p_{\pi_E}(s, a)}{p_{\pi}(s, a)}\right) \\
& = \log(1+\phi(s,a)).
\end{split}
\end{equation}

\label{eq:gail_reward}
\cref{eq:gail_reward} shows that the adversarial reward only depends on the ratio $\phi(s,a)=\frac{p_{\pi_E}(s, a)}{p_{\pi}(s, a)}$. Intuitively, $r_{gail}$ incentivizes the RL agent towards under-visited state-actions, where $\phi(s,a)>1$, and away from over-visited state-actions, where $\phi(s,a)<1$. When $\pi_E$ and $\pi$ match exactly, $r_{gail}$ converges to an indicator function for the support of $\pi_E$, since $\phi(s, a)=1 ~ \forall ~ (s, a)\in \supp(\pi_E)$~\citep{goodfellow2014generative}. In practice, the adversarial reward is unlikely to converge, as $p_{\pi_E}$ is estimated from a finite set of expert demonstrations. Instead, the adversarial reward continuously drives the agent to explore by evolving the reward landscape.

In practice, AIL faces several challenges, such as potential training instability associated with adversarial training. \citealt{wang2019red} demonstrated empirically that the adversarial reward could be unreliable in regions of the state-action space where the expert data is sparse, causing the agent to diverge from the intended behavior. When the agent policy is substantially different from the expert, the discriminator could differentiate them with high confidence. As a result, the agent receives tiny and uninformative reward, which causes significant slow down in training, a scenario similar to the vanishing gradient problem in GAN training~\citep{arjovsky2017towards}.

On the other hand, \citealt{kostrikov2018discriminator} demonstrated that the adversarial reward $-\log D(s, a)$ encodes an implicit survival bias: the non-negative reward may lead to sub-optimal behaviors in goal-oriented tasks where the agent learns to move around the goal to accumulate rewards, instead of completing the tasks. While the authors address the issue by introducing absorbing states, the solution requires additional RL signals from the environment, such as access to the time limit of an environment for detecting early termination of training episodes. In \cref{sec:lunar}, We demonstrate empirically that our proposed method mitigates the issue, and is able to imitate the expert more robustly.

\subsection{Support-weighted Adversarial Imitation Learning}\label{sec:sail}
We propose a novel reward function that unifies the adversarial reward with the score derived from support estimation of the expert policy.
\begin{align}
\begin{split}\label{eq:sail}
     & r_{sail}(s, a)  = r_{red}(s, a)\cdot \hat{r}_{gail}(s, a)\\
    &\quad\textrm{where} \quad \hat{r}_{gail}(s, a)  = 1-D(s, a) \in [0, 1]\\
\end{split}
\end{align}
%
SAIL leverages the exploration mechanism offered by the adversarial reward $\hat{r}_{gail}$, and weigh the adversarial reward with $r_{red}$, a score derived from support estimation. Intuitively, $r_{red}$ may be interpreted as a confidence estimate on the reliability of the adversarial reward, based on the availability of training data. This is particularly useful in our task context, when only limited number of expert demonstrations is available. As support estimation only requires expert demonstrations, our method requires no further assumptions than the underlying AIL method used.

We use a bounded reward $\hat{r}_{gail}$ instead of the typical $-\log D(s, a) \in[0, \infty]$. The modification allows $r_{red}$ and $\hat{r}_{gail}$ to have the same range and thus contribute equally to the reward function. For all experiments, we include the comparison between the two rewards, and show that the bounded one generally produces more robust policies. In the rest of paper, we denote SAIL with the bounded reward as \sailb{}, and SAIL with the log reward as SAIL. Similarly, we denote GAIL using the bounded reward as \gailb{}.

To improve training stability, SAIL constrains the RL agent to the estimated support of the expert policy, where $\hat{r}_{gail}$ provides a more reliable RL signal~\cite{wang2019red}. As $r_{red}$ tends to be very small (ideally zero) for $(s, a) \not\in \text{supp}(\pi_E)$, $r_{sail}$ discourages the agent from exploring those state-actions by masking away the rewards. This is a desirable property as the quality of the RL signals beyond the support of the expert policy can't be guaranteed. We demonstrate in \cref{sec:mujoco} the improved training stability on the Mujoco benchmark tasks .

SAIL also mitigates the survival bias in goal-oriented tasks by encouraging the agent to stop at the goal and complete the task. In particular, $r_{red}$ shapes the adversarial reward by favoring stopping at the goal against all other actions, as stopping at the goal is on the support of the expert policy, while other actions are not. We demonstrate empirically that SAIL learns to assign significantly higher reward towards completing the task and corrects for the bias in \cref{sec:lunar}. 

We provide the pseudocode implementation of SAIL in \cref{alg:sail}. The algorithm computes $r_{red}$ by estimating the support of the expert policy, followed by iterative updates of the policy and $\hat{r}_{gail}$. We apply the Trust Region Policy Optimization (TRPO) algorithm~\cite{schulman2015trust} with the reward $r_{sail}$ for policy updates.

\begin{algorithm}[tb]
   \caption{Support-weighted Adversarial Imitation Learning}
   \label{alg:sail}
\begin{algorithmic}
   \STATE {\bfseries Input:} Expert trajectories $\tau_E=\{(s_i, a_i)\}_{i=1}^N$, $\Theta$ function models, initial policy $\pi_{\omega_0}$, initial discriminator parameters $w_0$, learning rate $l_{D}$.
   \STATE
   \STATE $r_{red} = \text{RED}(\Theta, \tau_E)$
   \STATE {\bfseries for}~ $i=0, 1, \dots$
   \STATE\qquad sample a trajectory $\tau_i \sim \pi$
   \STATE\qquad $w_{i+1} = w_{i} + l_{D}~ (\hat{\E}_{\tau_i} (\triangledown\log D_{w_i}(s, a)) + \hat{\E}_{\tau_E}(\triangledown \log (1-D_{w_i}(s, a))))$
   \STATE\qquad $\hat{r}_{gail}: (s, a) \mapsto 1-D_{w_{i+1}}(s, a)$
   \STATE\qquad $\pi_{\omega_{i+1}} = $ {\sc TRPO}$(r_{red} \cdot \hat{r}_{gail},\pi_{\omega_i})$.
    \STATE
    \STATE{\bfseries def}~ {\sc RED}$(\Theta, \tau)$
    \STATE\qquad Sample $\theta\in\Theta$
   \STATE\qquad $\hat\theta=${\sc Minimize}$(f_{\hat{\theta}},f_{\theta},\tau)$
   \STATE\qquad {\bfseries return} {$r_{red}: (s, a) \mapsto \exp(-\sigma||f_{\hat{\theta}}(s, a) - f_{\theta}(s, a)||_2^2)$}
\end{algorithmic}
\end{algorithm}

\subsection{Comparing SAIL with GAIL}
\label{sec:sample_complexity}
In this section, we show that SAIL is at least as efficient as GAIL in its sample complexity for expert data, and provide comparable RL signals on the expert policy's support. We note that our analysis could be similarly applied to other AIL methods, suggesting the broad applicability of our approach.

We begin with the asymptotic setting, where the number of expert trajectories tends to infinity. In this case, both GAIL's, RED's and SAIL's discriminators ultimately recover the expert policy's support at convergence (see \cite{ho2016generative} for GAIL and \cite{wang2019red} for RED; SAIL follows from their combination). At convergence, both SAIL and GAIL also recovers the expert policy as the expert and agent policy distributions match exactly. It is therefore critical to characterize the rates of convergence of the two methods, namely their relative sample complexity with respect to the number of expert demonstrations. 

if the expert policy has infinite support, $r_{red}$ would converge to a constant function with value 1 under the asymptotic setting. We consequently recover GAIL and maintains all the theoretical properties of the algorithm. On the other hand, when only a finite number of demonstrations is available, $r_{red}$ would estimate a finite support and helps the RL agent to avoid state-actions not on the estimated support of the expert policy.

Formally, let $(s,a)\not\in\textnormal{supp}(\pi_E)$. Prototypical learning bounds for an estimator of the support $\hat r\geq0$ provide high probability bounds in the form of $\pp(\hat r(s,a) \leq  c \log(1/\delta) n^{-\alpha}) > 1-\delta$ for any confidence $\delta\in(0,1]$, with $c$ a constant not depending on $\delta$ or the number $n$ of samples (i.e., expert state-actions). Here, $\alpha>0$ represents the learning rate, namely how fast the estimator is converging to the support. By choosing the reward in \cref{eq:sail}, we are leveraging the faster learning rates between $\alpha_{red}$ and $\alpha_{gail}$, with respect to support estimation. At the time being, no results are available to characterize the sample complexity of GAIL (loosely speaking, the $\alpha$ and $c$ introduced above). Therefore, we proceed by focusing on a relative comparison with SAIL. In particular, we show the following (see appendix for a proof).

\begin{proposition}\label{prop:rates}
Assume that for any $(s,a)\not\in\supp(\pi_E)$ the rewards for RED and GAIL have the following learning rates in estimating the support
\begin{equation}\label{eq:assumption-rates}
\begin{split}
    & \pp\left(r_{red}(s,a)> \frac{c_{red}\log\frac{1}{\delta}}{n^{\alpha_{red}}} \right) \leq \delta \\
    & \pp\left(\hat{r}_{gail}(s,a)> \frac{c_{gail}\log\frac{1}{\delta}}{n^{\alpha_{gail}}} \right) \leq \delta.
\end{split}
\end{equation}
Then, for any $\delta\in(0,1]$ and any $(s,a)\not\in\supp(\pi_E)$, the following holds
\begin{equation}\label{eq:best-of-both}
    r_{sail}(s,a) \leq \min\left(\frac{c_{red}R_{gail}}{n^{\alpha_{red}}},\frac{c_{gail}R_{red}}{n^{\alpha_{gail}}}\right)\log\frac{1}{\delta},
\end{equation}
with probability at least $1-\delta$, where $R_{red}$ and $R_{gail}$ are the upper bounds for $r_{red}$ and $\hat{r}_{gail}$, respectively.
\end{proposition}

\cref{eq:best-of-both} shows that SAIL is at least as fast as the faster among RED and GAIL with respect to support estimation, implying that SAIL is at least as efficient as GAIL in the sample complexity for expert data. Moreover, \cref{eq:best-of-both} indicates how fast the proposed method could correctly identify state-actions not belonging to the estimated support of the expert and assigns low rewards to them.

\begin{proposition}\label{prop:comp}
For any $(s,a)\in\supp(\pi_E)$ and any $\delta\in(0,1]$, we assume that
\begin{equation}\label{eq:additional-assumption}
    \pp\left(|r_{red}(s,a)-1| > \frac{c_{red}\log\frac{1}{\delta}}{n^{\alpha_{red}}} \right) < \delta.
\end{equation}

The following event holds with probability at least $1-\delta$ that 
\begin{equation}\label{eq:almost-like-gail}
    \left|r_{sail}(s,a) - \hat{r}_{gail}(s,a)\right| \leq \frac{c_{red}R_{gail}}{n^{\alpha_{red}}}~\log\frac{1}{\delta}.
\end{equation}
\end{proposition}

\cref{eq:almost-like-gail} shows that on the expert policy's support, $r_{sail}$ is close to $\hat{r}_{gail}$ up to a precision that improves with the number of expert state-actions. SAIL thus provides RL signals comparable to GAIL on the expert policy's support.

It is also worth noting that the analysis could explain why $r_{red} + \hat{r}_{gail}$ is a less viable approach for combining the two RL signals. The analogous bound to \cref{eq:best-of-both} would be the sum of errors from the two methods, implying the {\em slower} of the two learning rates, while \cref{eq:almost-like-gail} would improve only by a constant, as $R_{gail}$ would be absent from \cref{eq:almost-like-gail}. Our preliminary experiments indicated that $r_{red} + \hat{r}_{gail}$ performed noticeably worse than \cref{eq:sail}.

Lastly, we comment on whether the assumptions in \cref{eq:assumption-rates,eq:additional-assumption} are satisfied in practice. Following the kernel-based version of RED~\citep{wang2019red}, we can borrow previous results from the set learning literature, which guarantee RED to have a rate of $\alpha_{red} = 1/2$~ \citep{de2014universally,rudi2017regularized}. These rates have been shown to be {\em optimal}. Any estimator of the support cannot have faster rates than $n^{-1/2}$, unless additional assumptions are imposed. Learning rates for distribution matching with GANs are still an active area of research, and conclusive results characterizing the convergence rates of these estimators are not available. We refer to \cite{singh2018nonparametric} for an in-depth analysis of the topic.


\section{Experiments}
\label{sec:exp}
To demonstrate improved performance and training stability, We evaluate SAIL against baseline methods on six Mujoco control tasks. In addition, we use Lunar Lander, another common benchmark task that allows easy access to human demonstrations, to show how the proposed method is more robust and mitigates reward bias. We omit evaluation against methods using off-policy RL algorithms, as they are not the focus of this work. As discussed previously, the proposed method could also be applied to such algorithms.

\subsection{Mujoco Tasks}
\label{sec:mujoco}
Mujoco control tasks have been commonly used as the standard benchmark for AIL. We evaluate SAIL against BC, GAIL and RED on Hopper, Reacher, HalfCheetah, Walker2d, Ant and Humanoid. We adopt the same experimental setup presented in \cite{ho2016generative} by sub-sampling the expert trajectories every 20 samples. Consistent with the observation from \cite{kostrikov2018discriminator}, our preliminary experiments show that sub-sampling presents a more challenging setting, as BC is competitive with AIL when full trajectories are used. In our experiments, we also adopt the minimum number of expert trajectories specified in \citealt{ho2016generative} for each task. More details on experiment setup are available in the appendix.

We run each algorithm using 5 different random seeds in all Mujoco tasks. \cref{tab:comp} shows the performances among the evaluated algorithms. We choose the best policy obtained from the 5 random seeds for each algorithm, and report the mean performance and standard deviation of the chosen policy over 50 evaluation runs. The policies are rolled out deterministically.

\begin{table}[tbh]
\caption{Episodic reward on the Mujoco tasks evaluated over 50 runs. \sailb{} achieves overall the best performance, with significantly lower standard deviation, indicating the robustness of the learned policies.}
\label{tab:comp}
\vskip 0.15in
\begin{center}
\begin{small}
\begin{sc}
\noindent\setlength\tabcolsep{3pt}
\setlength{\extrarowheight}{1pt}
\begin{tabular}{lllllll}
\toprule
& Hopper& Reacher& Cheetah& Walker& Ant& Humanoid\\
\midrule
BC & 312.3 $\pm$ 34.5& -8.8 $\pm$ 3.3& 1892.0 $\pm$ 206.9& 248.2 $\pm$ 117.8& 1752.0 $\pm$ 434.8& 539.4 $\pm$ 185.7\\
RED & 1056.5 $\pm$ 0.5& -9.1 $\pm$ 4.1& -0.2 $\pm$ 0.7& 2372.8 $\pm$ 8.8& 1005.5 $\pm$ 8.6& 6012.0 $\pm$ 434.9\\
GAIL & \bf{3826.5 $\pm$ 3.2}& -9.1 $\pm$ 4.4& 4604.7 $\pm$ 77.6& 5295.4 $\pm$ 44.1& 1013.3 $\pm$ 16.0& 8781.2 $\pm$ 3112.6\\
\gailb{} & 3810.5 $\pm$ 8.1& -8.3 $\pm$ 2.5& 4510.0 $\pm$ 68.0& 5388.1 $\pm$ 161.2& 3413.1 $\pm$ 744.7& 10132.5 $\pm$ 1859.3\\
SAIL & 3824.7 $\pm$ 6.6 & -7.5 $\pm$ 2.7 & \bf{4747.5 $\pm$ 43.4} & 5293.0 $\pm$ 590.9& 3330.4 $\pm$ 729.4& 9292.8 $\pm$ 3190.0\\
\sailb{} & 3811.6 $\pm$ 3.8& \bf{-7.4 $\pm$ 2.5} & 4632.2 $\pm$ 59.1& \bf{5438.6 $\pm$ 18.4} & \bf{4176.3 $\pm$ 203.1} & \bf{10589.6 $\pm$ 52.2}\\
\bottomrule
\end{tabular}
\end{sc}
\end{small}
\end{center}
\end{table}

The results show that \sailb{} is comparable to GAIL on Hopper, and outperform the other methods on all other tasks. We note that RED significantly underperforms in the sub-sampling setting\footnote{\citealt{wang2019red} used full trajectories in their original experiments}. Across all tasks, \sailb{} generally achieves lower standard deviation compared to other algorithms, in particular for Humanoid, indicating the robustness of the learned policies.

We stress that standard deviation is a critical metric for the robustness of the learned policies and has practical implications. For instance, the large standard deviations in Humanoid are caused by occasional crashes, which is potentially dangerous in real-world applications regardless of the general good performance. To illustrate this, \cref{fig:humanoid_hist} shows the histogram of all 50 evaluations in Humanoid for RED, \gailb{} and \sailb{}. It is clear that \sailb{} imitates the expert consistently. Though \gailb{} appears to be only slightly worse in average performance, the degradation is caused by occasional and highly undesirable crashes, suggesting incomplete imitation of the expert. RED learns a sub-optimal gait, but demonstrates no crashes. The results suggest that the proposed method improves the quality of the RL signals.

\begin{figure}[tbh]
    \centering   
    \begin{subfigure}[b]{.32\textwidth}
        \includegraphics[width=\textwidth, trim={0 0 1.5cm 0.5 cm}]{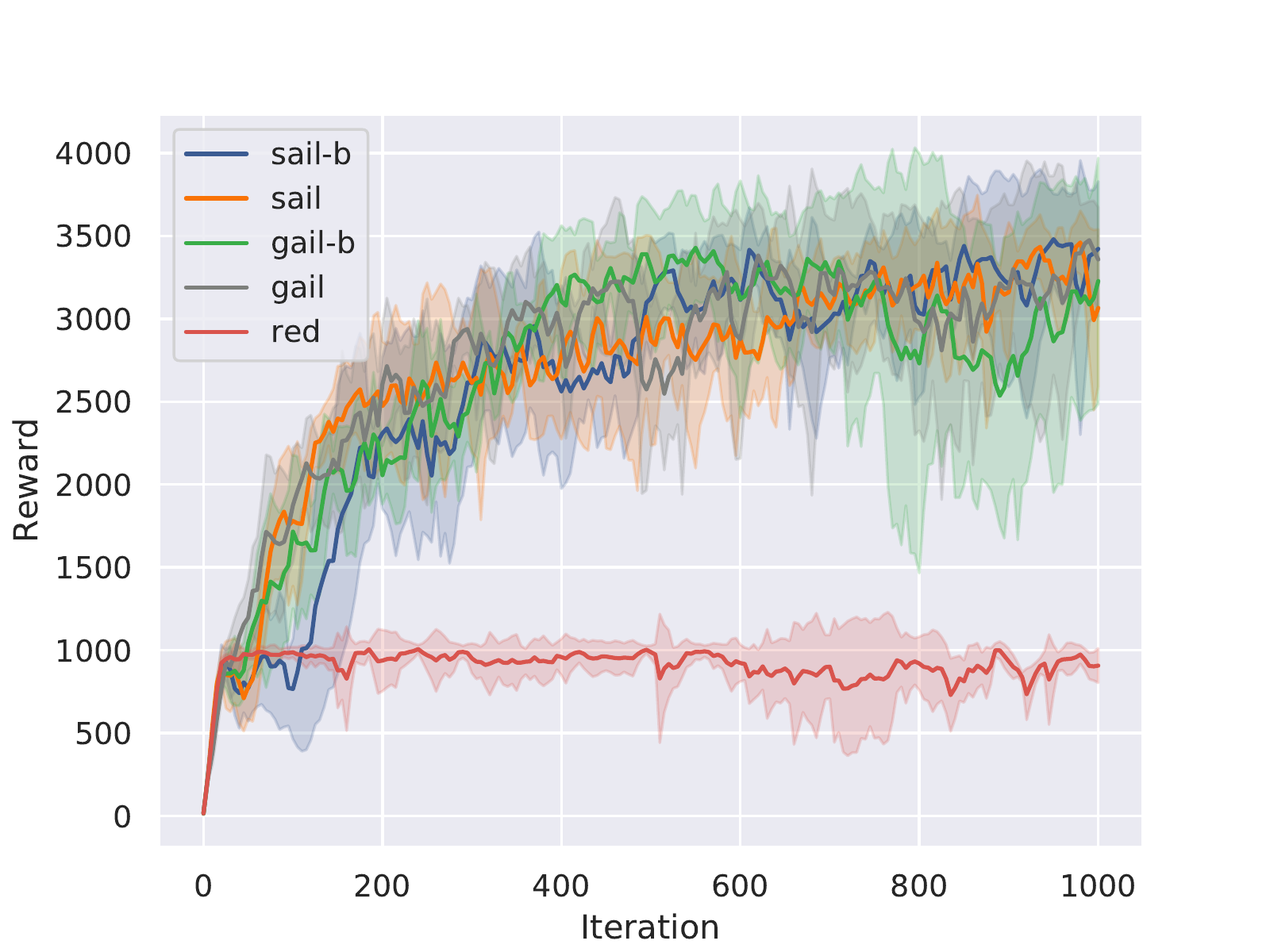}
        \caption{Hopper}
        \label{fig:Hopper}
    \end{subfigure}
    \begin{subfigure}[b]{.32\textwidth}
        \includegraphics[width=\textwidth, trim={0 0 1.5cm 0.5 cm}]{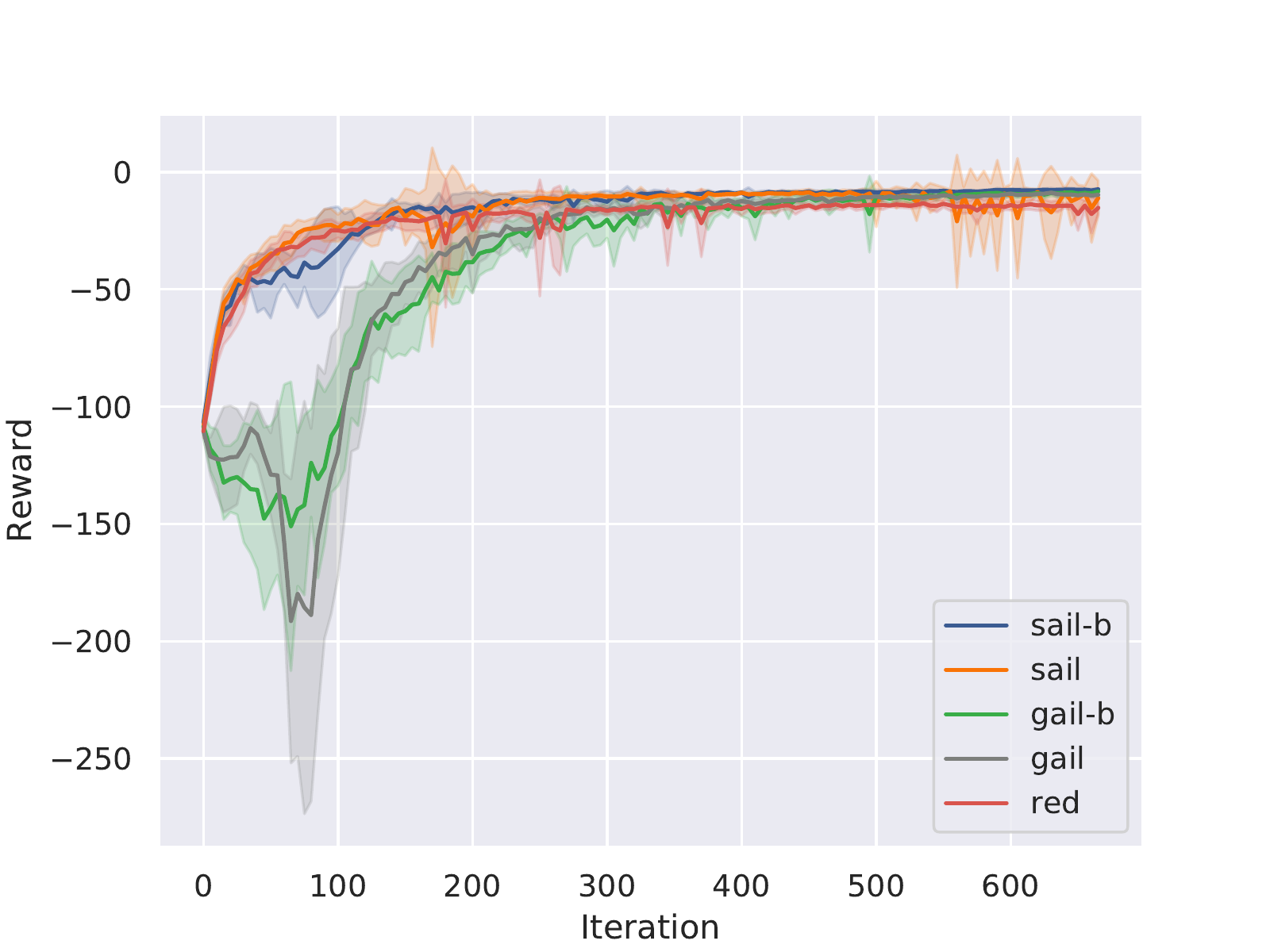}
        \caption{Reacher}
        \label{fig:Reacher}
    \end{subfigure}
    \begin{subfigure}[b]{.32\textwidth}
        \includegraphics[width=\textwidth, trim={0 0 1.5cm 0.5 cm}]{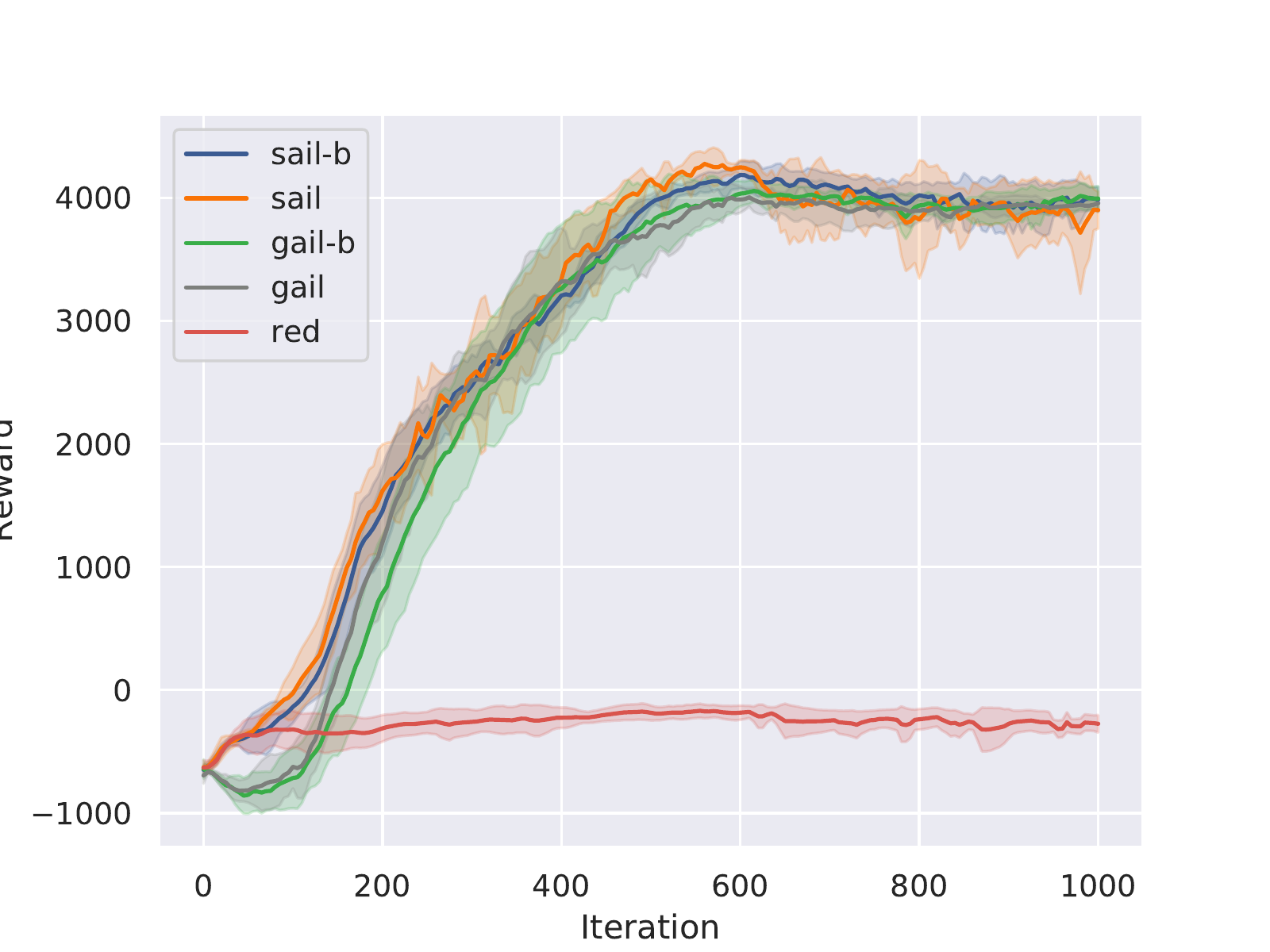}
        \caption{HalfCheetah}
        \label{fig:HalfCheetah}
    \end{subfigure}
    
    \begin{subfigure}[b]{.32\textwidth}
        \includegraphics[width=\textwidth, trim={0 0 1.5cm 0.5 cm}]{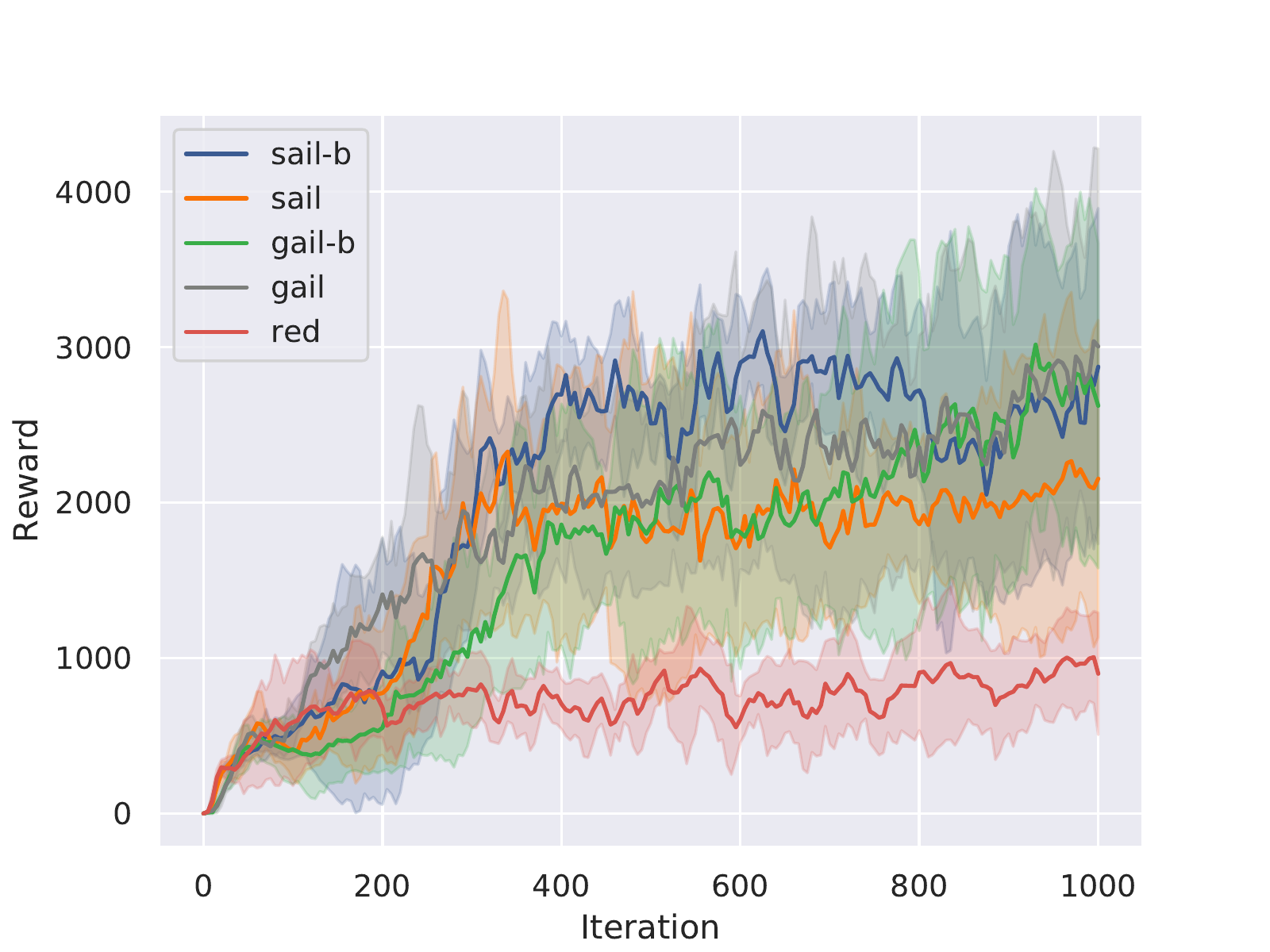}
        \caption{Walker2d}
        \label{fig:Walker2d}
    \end{subfigure}
    \begin{subfigure}[b]{.32\textwidth}
        \includegraphics[width=\textwidth, trim={0 0 1.5cm 0.5 cm}]{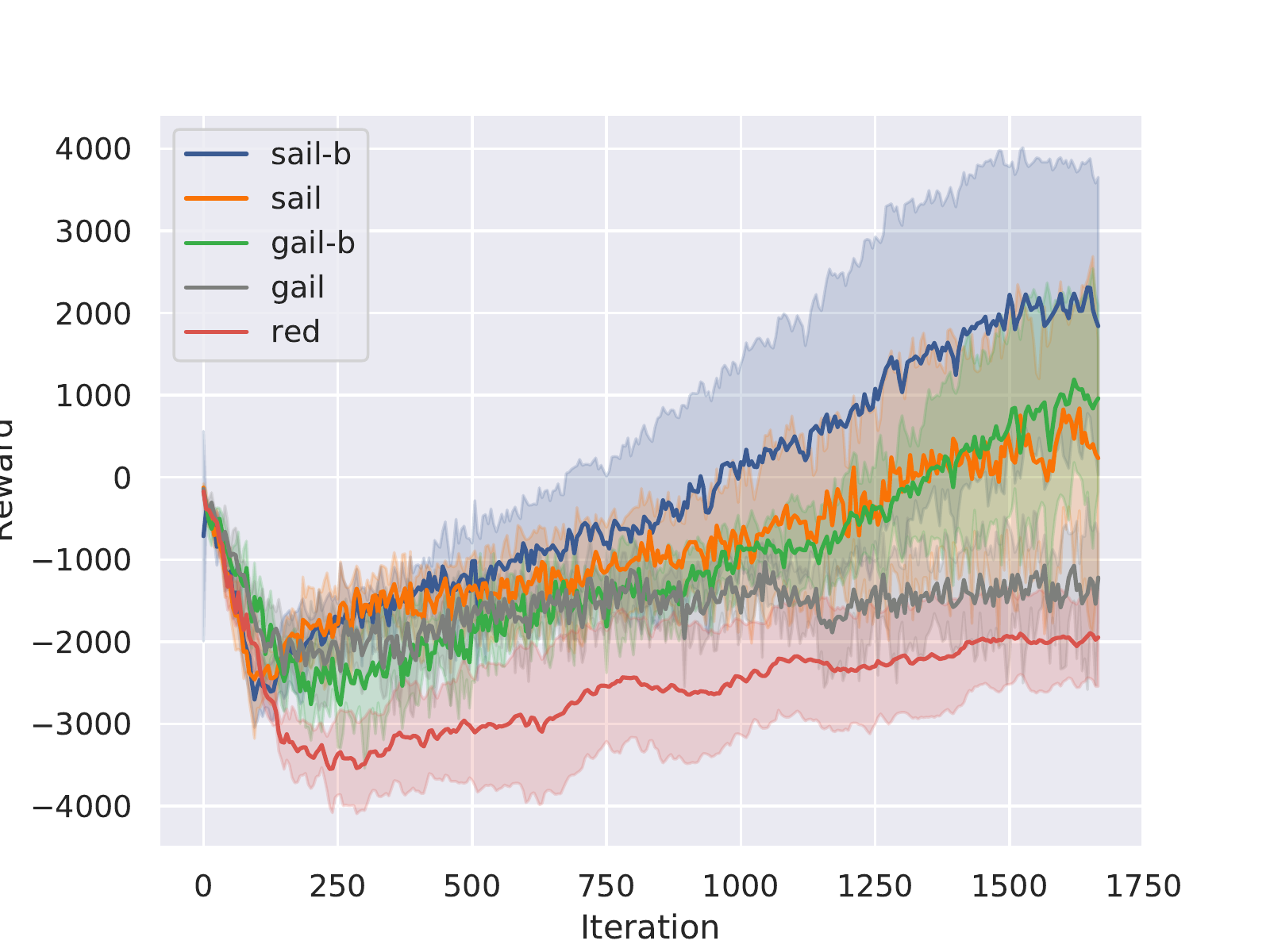}
        \caption{Ant}
        \label{fig:Ant}
    \end{subfigure}
    \begin{subfigure}[b]{.32\textwidth}
        \includegraphics[width=\textwidth, trim={0 0 1.5cm 0.5 cm}]{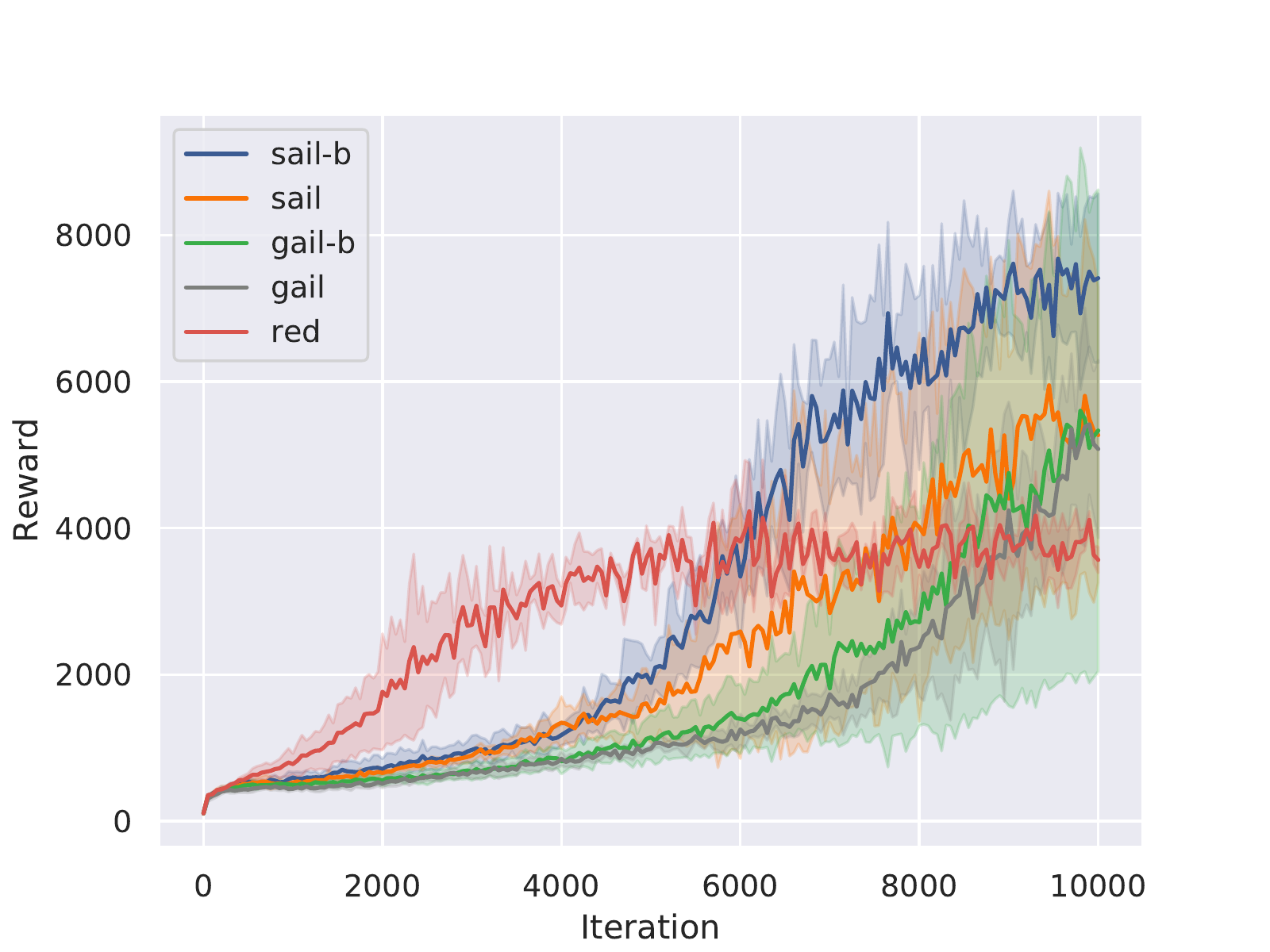}
        \caption{Humanoid}
        \label{fig:Humanoid}
    \end{subfigure}
    \caption{Training progress for RED, GAIL, \gailb{}, SAIL, and \sailb{}. Consistent with our theoretical analysis, \sailb{} (blue) is more stable and sample efficient in Reacher, Ant and Humanoid, and comparable to other algorithms for the remaining tasks.}\label{fig:train_plot}
\end{figure}

For ablation, we compare the unbounded reward (i.e. SAIL) against against the bounded one (i.e. \sailb{}), we observe that the bounded variant generally produces policies with smaller standard deviations and better performances, especially for Ant and Humanoid. We attribute the improvements to the fact that \sailb{} receives equal contribution in RL signal from both support estimation and adversarial reward, as $r_{red}$ and $\hat{r}_{gail}$ have the same range. We also note that GAIL fails to imitate the expert in Ant, while \gailb{} performs significantly better. The results suggest that constraining the range of the adversarial reward could improve performance.

\subsubsection{Training Stability and Sample Efficiency}
\label{sec:efficient}
To assess algorithms' respective sensitivity to random seeds, we plot the policy performance against the number of iterations for each algorithm in \cref{fig:train_plot}, Each iteration consists of 1000 environment steps. The figure reports mean and standard deviation of each algorithm, across the 5 random seeds.

\cref{fig:train_plot} shows that \sailb{} is more sample efficient and stable in Reacher, Ant and Humanoid tasks; and is comparable to the other algorithms in the remaining tasks. Consistent with our analysis in \cref{sec:sample_complexity}, \sailb{} appears at least as efficient as GAIL even when the support estimation (i.e., the performance of RED) suffers from insufficient expert data in Hopper, HalfCheetah and Walker2d. In Reacher, Ant and Humanoid, \sailb{} benefits from the support estimation and achieves better performance and training stability. In particular, we note that without support estimation, GAIL fails to imitate the expert in Ant (\cref{fig:Ant}). Similar failures were also observed in \cite{kostrikov2018discriminator}. GAIL is also more sensitive to initial conditions: GAIL converged to sub-optimal policies in 2 out 5 seeds in Humanoid. Lastly, while RED improves noticeably faster during early training in Humanoid, it converged to a sub-optimal behavior eventually.

\subsection{Lunar Lander}
\label{sec:lunar}
\begin{figure}[tbh]
    \centering
    \begin{subfigure}{0.46\textwidth}
        \caption{Average environment reward and standard deviation on Lunar Lander, evaluated over 50 runs for the default and no-terminal environment.}
        \label{tab:lunar}
        \begin{small}
        \begin{sc}
        \begin{tabular}{lll}
        \toprule
        & Default & No-terminal \\
        \midrule
        BC & 100.38 $\pm$ 130.91 & 100.38 $\pm$ 130.91\\
        RED & 13.75 $\pm$ 53.43 & -39.33 $\pm$ 24.39\\
        GAIL & 258.30 $\pm$ 28.98 & 169.73 $\pm$ 80.84\\
        \gailb{} & 250.53 $\pm$ 67.07 & -69.33 $\pm$ 79.76\\
        SAIL & 257.02 $\pm$ 20.66 & 237.96 $\pm$ 49.70\\
        \sailb{} & \bf{262.97 $\pm$ 18.11} & \bf{256.83 $\pm$ 20.99}\\
        Expert & 253.58 $\pm$ 31.27 & 253.58 $\pm$ 31.27\\
        \bottomrule
        \end{tabular}
        \end{sc}
        \end{small}
    \end{subfigure}
    \quad
    \begin{subfigure}{0.46\textwidth}
        \centering
        \includegraphics[width=0.86\textwidth, trim={0.2cm 0.22cm 0.22cm 0.22cm}]{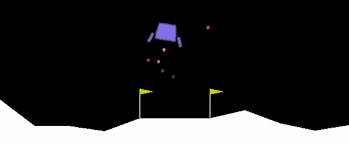}
        \caption{The task of Lunar Lander requires landing the spacecraft between the flags without crashing.}%
        \label{fig:lunar}
    \end{subfigure}
    \caption{}
\end{figure}

We demonstrate that the proposed method mitigates the survival bias in Lunar Lander (\cref{fig:lunar}) from OpenAI Gym~\citep{gym}, while other baseline methods imitate the expert inconsistently. In this task, the agent is required to control a spacecraft to safely land between the flags. We specifically choose a human expert to provide 10 demonstrations as 
alternative source for training data, as the task allows easy access to human demonstrations. This is in contrast with Mujoco tasks, where human demonstration is difficult and the expert policies are learned via RL.

\begin{figure}[tbh]
    \begin{center}
    \begin{subfigure}{0.46\textwidth}
        \includegraphics[width=0.9\textwidth]{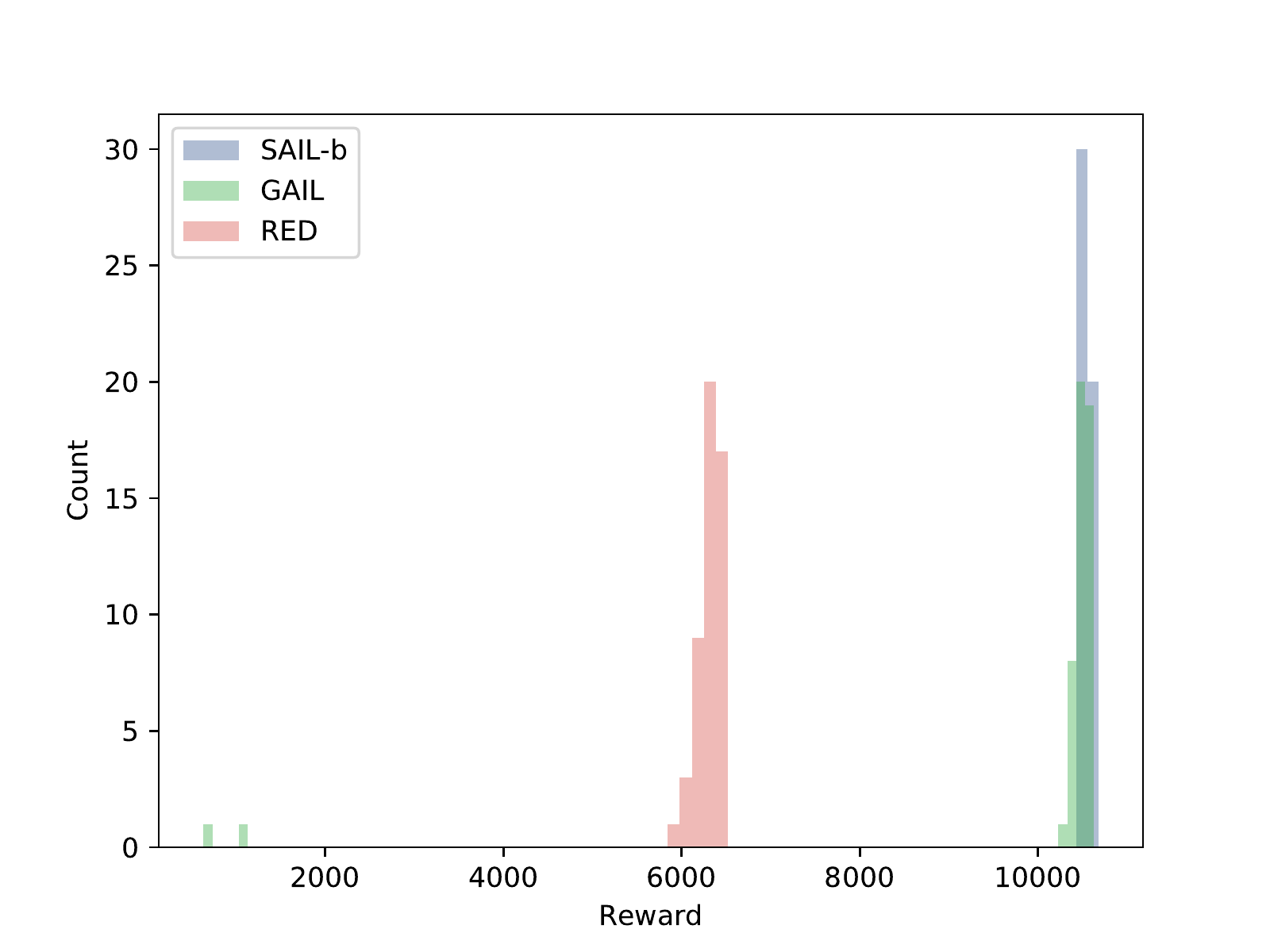}%
        \caption{
          Performance histogram of 50 evaluation runs on Humanoid for RED, GAIL, and \sailb{}. \sailb{} imitates the expert consistently. GAIL has undesirable failure cases, with rewards of less than 1000 (bottom left corner). RED is consistent though sub-optimal.}
        \label{fig:humanoid_hist}
    \end{subfigure}
    \qquad
    \begin{subfigure}{0.46\textwidth}
       \includegraphics[width=0.98\textwidth]{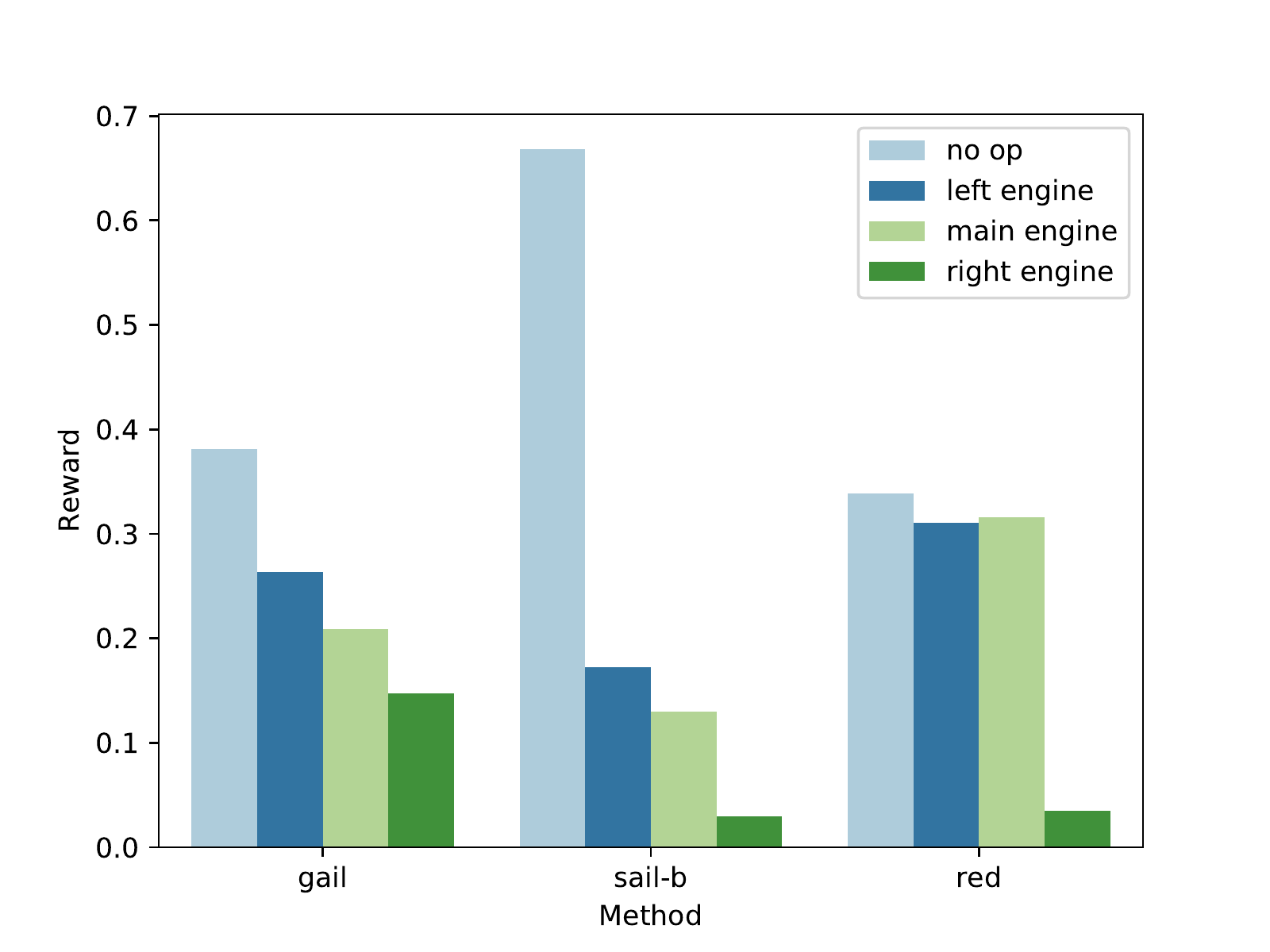}
        \caption{(LunarLander) Average reward at the goal states by different algorithms. \sailb{} assigns significantly higher reward to "no op", enabling the agent to learn the appropriate landing behaviors. Other algorithms fail to imitate the expert consistently.}
    \label{fig:lunar_reward}
    \end{subfigure}
    \end{center}
    \caption{}
\end{figure}

We observe that even without the environment reward, Lunar Lander provides a natural RL signal by terminating episodes early after a crash, thus encouraging the agent to avoid crashing. Consequently, both SAIL and GAIL are able to successfully imitate the expert and land the spacecraft appropriately. SAIL performs slightly better than GAIL on the average reward, and achieve noticeably lower standard deviation. The average performances and the standard deviations evaluated over 50 runs are presented in \cref{tab:lunar}.

To construct a more challenging task, we disable early termination from environment, thus removing external RL signals. In this no-terminal environment, a training episode only ends after the time limit. We present each algorithm's performance for the no-terminal setting in \cref{tab:lunar}. SAIL outperforms GAIL. Specifically, we observe that GAIL learns to land for some initial conditions, while exhibiting survival bias in other scenarios by hovering at the goal. In contrast, SAIL is still able to recover the expert policy.\footnote{Illustrative video at \url{https://vimeo.com/361835881}}

We show in \cref{fig:lunar_reward} the average learned reward fr GAIL, \sailb{} and RED at goal states, to visualize how support estimation shapes the learned reward. The goal states are selected from the expert trajectories and satisfy two conditions: $1)$ touching the ground (the state vector has indicator variables for ground contact), and $2)$ has "no op" as the corresponding action. As the reward functions are dynamic, we snapshot the learned rewards when the algorithms obtain their best policies, respectively. It is clear that \sailb{} assigns a significantly higher average reward to "no op" at goal states compared against the other algorithms, thus facilitating the agent learning. Though GAIL and RED still favor "no op" to other actions, the differences in reward are much smaller, causing less consistent landing behaviors.

We further observe that all evaluated AIL methods oscillate between partially hovering behavior and landing behavior during policy learning. The observation suggests that our method only partially addresses the survival bias, a limitation we will tackle in future works. This is likely caused by SAIL's non-negative reward, despite the beneficial shaping effect from support estimation.

To demonstrate the compatibility of the proposed method with other improvements for AIL, we show that the method is compatible with the absorbing state technique proposed in \citealt{kostrikov2018discriminator} if the time limit of an environment is known. The additional experiment results and discussion are available in the Appendix.

\section{Conclusion}
\label{sec:con}
In this paper, we propose Support-weighted Adversarial Imitation Learning by combining support estimation of the expert policy with adversarial imitation learning. The proposed approach improves the quality of the RL signals by weighing the adversarial reward with the score derived from support estimation, leading to better training stability and performance. Our approach is also orthogonal and complementary to many existing AIL methods. More broadly, our results show that expert demonstrations contain rich sources of information for imitation learning. Effectively merging different sources of information in the expert demonstrations produces more efficient and stable algorithms; and appears to be a promising direction for future research.


\bibliography{main.bib}
\bibliographystyle{icml2020}

\appendix
\newpage
\onecolumn
\section{Proof for Proposition \ref{prop:rates} and \ref{prop:comp}}
\label{sec:app-rates}
Observe that for any $(s,a)\in S \times A$
\begin{equation}\label{eq:rsail-less-than-min-and-upper-bound}
    r_{sail}(s,a) = r_{red}(s,a)\cdot r_{gail}(s,a) \leq \min(r_{red}(s,a)R_{gail},r_{gail}(s,a)R_{red}). 
\end{equation}
By the assumption on the learning rate in \cref{eq:assumption-rates}, one of the two following events holds with probability at least $1-\delta$, for any $(s,a)\not\in\textnormal{supp}(\pi_E)$ and $\delta\in(0,1]$
\begin{equation}
    r_{red}(s,a) \leq \frac{c_{red}\log\frac{1}{\delta}}{n^{\alpha_{red}}} \qquad \text{or} \qquad r_{gail}(s,a) \leq \frac{c_{gail}\log\frac{1}{\delta}}{n^{\alpha_{gail}}}.
\end{equation}
Plugging the above upper bounds into \cref{eq:rsail-less-than-min-and-upper-bound} yields the desired result in \cref{eq:best-of-both}.

By assumption in \cref{eq:additional-assumption} the following event holds with  probability at least $1-\delta$ for $(s,a)\in\textnormal{supp}(\pi_E)$. 
\begin{equation}
    |r_{red}(s,a) - 1| \leq \frac{c_{red}\log\frac{1}{\delta}}{n^{\alpha_{red}}}.
\end{equation}
Plugging this inequality in the definition of $r_{sail}$, we obtain
\begin{align}
    \pushQED{\qed}     
    |r_{sail}(s,a) - r_{gail}(s,a)| & = |r_{gail}(s,a)(r_{red}(s,a) - 1)|\\
    & \leq |r_{gail}(s,a)||r_{red}(s,a) - 1|\\
    & \leq \frac{c_{red}R_{gail}\log\frac{1}{\delta}}{n^{\alpha_{red}}},
\end{align}
\qedhere
\popQED

\section{Experiment Details}
The experiments are based on OpenAI's baselines\footnote{\url{https://github.com/openai/baselines}} and the original implementation of RED\footnote{\label{code:red}\url{https://github.com/RuohanW/RED}}. We adapted the code from RED\footnoteref{code:red} for our experiments, and used the accompanying dataset of expert trajectories. 4 Nvidia GTX1070 GPUs were used in the experiments.

\cref{tab:env_info} shows the environment information, number of environment steps and number of expert trajectories used for each task. Each full trajectory consists of 1000 $(s, a)$ pairs. They are sub-sampled during the experiments.

\begin{table}
  \centering
  \begin{tabular}{llllll}
    Task & State Space & Action Space & Trajectories & Env Steps & Exp Performance\\
    \midrule
    Hopper-v2 & 11 & 3 & 4  & $3\times 10^6$ & 3777.8 $\pm$ 3.8\\
    Reacher-v2 & 11 & 2 & 4 & $1\times 10^6$ & -3.7 $\pm$ 1.4     \\
    HalfCheetah-v2 & 17 & 6 & 4  & $3\times 10^6$ & 4159.8 $\pm$ 93.1 \\
    Walker2d-v2 & 17 & 6 & 4  & $3\times 10^6$ & 5505.8 $\pm$ 81.4   \\
    Ant-v2 & 111 & 8 & 4  & $5\times 10^6$ & 4821.0 $\pm$ 107.4    \\
    Humanoid-v2 & 376 & 17 & 80 & $3\times 10^7$ & 10413.1 $\pm$ 47.0\\
    \bottomrule
  \end{tabular}
  \vskip 0.2in
  \caption{Environment information, number of expert trajectories and environment steps used for each task}
  \label{tab:env_info}
\end{table}

\subsection{Network Architecture}
The default policy network from OpenAI's baselines are used for all tasks: two fully-connected layers of 100 units each, with tanh nonlinearities. The discriminator networks and the value function networks use the same architecture.

RED and SAIL use RND~\cite{burda2018exploration} for support estimation. We use the default networks from RED\footnoteref{code:red}. We set $\sigma$ following the heuristic in \cite{wang2019red} that $(s, a)$ from the expert trajectories mostly have reward close to 1.

\subsection{Hyperparameters}
For fair comparisons, all algorithms shared hyperparameters for each task. We present them in the table below, including discriminator learning rate $l_D$, discount factor $\gamma$, number of policy steps per iteration $n_G$, and whether the policy has fixed variance. All other hyperparameters are set to their default values from OpenAI's baselines.

\begin{table}
  \centering
  \begin{tabular}{ccccc}
    Task Name & $\gamma$ & $l_D$ & $n_G$ & Fixed Variance \\
    \midrule
    Hopper & 0.99  & 0.0003 & 3 & False \\
    Reacher & 0.99  & 0.0003 & 3 & False \\
    HalfCheetah & 0.99  & 0.0003 & 3 & False \\
    Walker2d & 0.99  & 0.0003 & 3 & False \\
    Ant & 0.99  & 0.0001 & 3 & False \\
    Humanoid & 0.99  & 0.0001 & 5 & False \\
    \bottomrule
  \end{tabular}
  \vskip 0.2in
  \caption{Hyperparameters used for each tasks}
  \label{tab:red_hyper}
\end{table}

\section{Additional Results on Lunar Lander}
In the default environment, Lunar Lander contains several terminal states, including crashing, flying out of view, and landing at the goal. In the \textit{no-terminal} environment, all terminal states are disabled, such that the agent must solely rely on the expert demonstrations for inferring that stopping is the correct behavior upon landing.

To compare our method with the technique of introducing virtual absorbing state (AS)~\citep{kostrikov2018discriminator}, we also construct a \textit{goal-terminal} environment where the only terminal state is successful landing at the goal, because the AS technique cannot be directly applied in the no-terminal environment. We also combine SAIL with the AS technique to demonstrate that the proposed method is complementary is existing improvements to AIL. We present the results in \cref{tab:as_cmp}.

\begin{table}[hb]
\label{tab:as_cmp}
\begin{tabular}{llll}
  \toprule
  & Default & Goal-terminal & No-terminal \\
  \midrule
  GAIL & 258.30 $\pm$ 28.98 & -7.16 $\pm$ 31.64 & -69.33 $\pm$ 79.76\\
  \gailb{} & 250.53 $\pm$ 67.07 & 4.16 $\pm$ 107.37 & 169.73 $\pm$ 80.84\\
  SAIL & 257.02 $\pm$ 20.66 & 261.07 $\pm$ 35.66 & 237.96 $\pm$ 49.70\\
  \sailb{} & 262.97 $\pm$ 18.11 & 252.07 $\pm$ 67.22 & 256.83 $\pm$ 20.99\\

  \midrule
  GAIL + AS & 271.46 $\pm$  11.90 & 110.22 $\pm$ 119.25 & -\\
  \gailb{} + AS & 269.97 $\pm$  16.48 & 186.02 $\pm$ 98.27 & -\\
  SAIL + AS & 274.89 $\pm$ 12.82 & 254.58 $\pm$  25.40 & -\\
  \sailb{} + AS & 270.33 $\pm$ 15.86 & 258.30 $\pm$ 20.75 & -\\
  \bottomrule
  \end{tabular}
  \caption{Average environment reward and standard deviation on Lunar Lander, evaluated over 50 runs for the default, goal-terminal and no-terminal environment.}
\end{table}

\end{document}